\documentclass[sigconf]{acmart}
\AtBeginDocument{%
  \providecommand\BibTeX{{%
    \normalfont B\kern-0.5em{\scshape i\kern-0.25em b}\kern-0.8em\TeX}}}

\acmYear{2021}\copyrightyear{2021}
\setcopyright{acmcopyright}
\acmConference[Destion '21]{Design Automation for CPS and IoT}{May 18, 2021}{Nashville, TN, USA}
\acmBooktitle{Design Automation for CPS and IoT (Destion '21), May 18, 2021, Nashville, TN, USA}
\acmPrice{15.00}
\acmDOI{10.1145/3445034.3460509}
\acmISBN{978-1-4503-8316-5/21/05}

\usepackage{algorithm2e}

\begin{document}

\title{Embedded Out-of-Distribution Detection on an Autonomous Robot Platform}

\author{Michael Yuhas, Yeli Feng, Daniel Jun Xian Ng, Zahra Rahiminasab, Arvind Easwaran}
\email{{michaelj004, yeli.feng, danielngjj, rahi0004,  arvinde}@ntu.edu.sg}
\affiliation{%
  \institution{Nanyang Technological University}
  \country{Singapore}
}

\renewcommand{\shortauthors}{Michael Yuhas, et al.}

\begin{abstract}
Machine learning (ML) is actively finding its way into modern cyber-physical systems (CPS), many of which are safety-critical real-time systems. It is well known that ML outputs are not reliable when testing data are novel with regards to model training and validation data, i.e., out-of-distribution (OOD) test data. We implement an unsupervised deep neural network-based OOD detector on a real-time embedded autonomous Duckiebot and evaluate detection performance. Our OOD detector produces a success rate of $87.5\%$ for emergency stopping a Duckiebot on a braking test bed we designed. We also provide case analysis on computing resource challenges specific to the Robot Operating System (ROS) middleware on the Duckiebot. 
\end{abstract}



\maketitle

\section{Introduction}
Machine learning's prediction power degrades when testing data shift away from distributions of model training data's underlying properties  \cite{varshney2017}. Such distribution shift may yield unsafe results.  As machine learning (ML) is rapidly entering modern cyber-physical systems (CPS), there is an urgency to assure that the uncertainty property of machine learning will not jeopardize the safety requirement of CPS. One emerging approach is to proactively detect whether distribution shift occurs in real-time input data and devise a fail-safe mechanism to prevent CPS, for example, an ML-enabled autonomous driving module, from acting on unreliable ML outputs. Along this line, many out-of-distribution (OOD) detection methods~\cite{vernekar2019out,bulusu2020anomalous,gu2019towards} have been proposed in the literature modeling from a supervised or unsupervised learning principle.

While OOD detection applies to all ML systems, CPS such as autonomous vehicles and robots pose unique challenges to algorithm efficiency due to real-time and embedded system constraints. In this paper, we study the performance of an existing OOD detection algorithm~\cite{ramakrishna2020efficient} on a real-time embedded Robot Operating System (ROS) based architecture; specifically the Duckietown robotics platform~\cite{duckietown}.

Contributions of our work include:
\begin{itemize}
\item We implemented and demonstrate that~\cite{ ramakrishna2020efficient} achieves efficient OOD detection in a system design with three ROS nodes. They include a variational autoencoder (VAE) based OOD detection node, an OpenCV-based lane following node and a motor steering node.
\item We designed an emergency braking system and show that its OOD detection accuracy is satisfactory. We suppose safe braking is a function of distance to a roadblock, robot speed, and classification threshold of the OOD detector; we provide case analysis to understand the limitations of the present design system parameters and discuss their impact on the emergency braking system's performance.
\end{itemize}

To the best of our knowledge, this is a pioneer work that explores the real-time implementation of efficient OOD detection on embedded platforms. The specific setup gives us very tight computing resources and a short detection time window due to the speed of Duckiebots.

The rest of the paper is organized into four sections. Section~\ref{sec:ood} provides a brief introduction of OOD detection followed by the detection method for experiments. Section~\ref{sec:framework} describes the Duckietown platform and the implementation we used for investigations. Section~\ref{sec:experiment} describes our experimental methodology, and Section~\ref{sec:results} presents results and case studies.

\section{Out-of-Distribution Detection}
\label{sec:ood}

The reliability of ML outputs is closely related to model training data. As a result, ML models that achieved high accuracy during training can produce high errors during testing, especially for novel samples. In autonomous driving, solutions that attempt to address this problem tend to introduce an independent detector to check test samples at run-time and flag warning signals if incoming samples shift away from distributions of model training data. 

In this paper, we focus on implementing and evaluating the algorithm efficiency of a $\beta$-VAE based OOD detection method proposed by~\cite{ramakrishna2020efficient}. We deploy the implementation on an embedded robot platform: the Duckietown framework.

Taking advantage of the information bottleneck of VAE \cite{kingma2013auto}, ~\cite{ramakrishna2020efficient} proposed to learn the underlying in-distribution properties of training data in an unsupervised way. A VAE has encoder and decoder parts. The encoder maps the input data to an encoded representation in the latent space, and the decoder reconstructs the image by sampling an encoded representation from the latent space. The VAE's learning objective is to minimize the distribution discrepancy between an unknown true prior $p(z)$ of the input space and approximated posterior $q(z \mid x)$ from the encoder output, and the reconstruction error between the decoder output and input. The approximated posterior represents the latent space distribution, which usually has a much lower dimension than the input data. ~\cite{ramakrishna2020efficient} proposed to use the distribution discrepancy in the latent space, specifically Kullback–Leibler (KL) divergence, to construct an OOD detection measure.

Let $x$ be in-distribution training images. The KL divergence between the approximated posterior of latent variable $z$ and the true prior will be minimized in a properly trained VAE model. Given OOD images, the encoder will return high KL values. We define OOD scores by equation \ref{eq:eq1}. 
\begin{equation}
\label{eq:eq1}
OODScore = \sum_{i\in\mathbb{Z}}D_{KL}(q(z|x) \mid p(z))
\end{equation} 

$\mathbb{Z}$ is a latent sub-space that encodes most information about OOD factors, hence enhances OOD detection performance.~\cite{ ramakrishna2020efficient} also applied a modified VAE, $\beta$-VAE~\cite{higgins2016beta}, which introduces a hyperparameter $\beta$ to further regulate the latent space distribution discrepancy. A carefully chosen $\beta$ value encourages the discovery of disentangled latent factors, further enhancing OOD detection performance.

\section{Duckietown Framework}
\label{sec:framework}

Duckietown is a low-cost, open-source platform targeting students and researchers interested in autonomous driving. The platform contains both hardware and software components; the hardware is designed using off-the-shelf components, and the software takes advantage of the modularity of ROS~\cite{ros_vol2} and Docker~\cite{docker} to provide a layer of abstraction between the sensors, actuators and any application running on the platform. In addition the Duckietown platform contains specifications for maps, obstacles, and world objects, but we do not exploit these in the scope of this paper~\cite{duckietown}.

Furthermore, Duckietown has already been used in previous research on ML and CPS. The Duckietown gym, a simulation environment provided as part of the Duckietown platform, was used to demonstrate reinforcement learning~\cite{9259236}. ML models were first trained in the Duckietown gym and then used to control a physical Duckiebot in~\cite{9263751,9263406,9207497}. These experiments demonstrate the potential of using Duckietown as a test bed for the evaluation of OOD detection algorithms outside of a simulated environment without the need for costly hardware. In addition,  several autonomous car development frameworks such as Autoware~\cite{kato2015open} are also based on ROS, which implies that the software abstraction layer of Duckietown is also representative.

\begin{figure}
  \centering
  \includegraphics[width=0.6\linewidth]{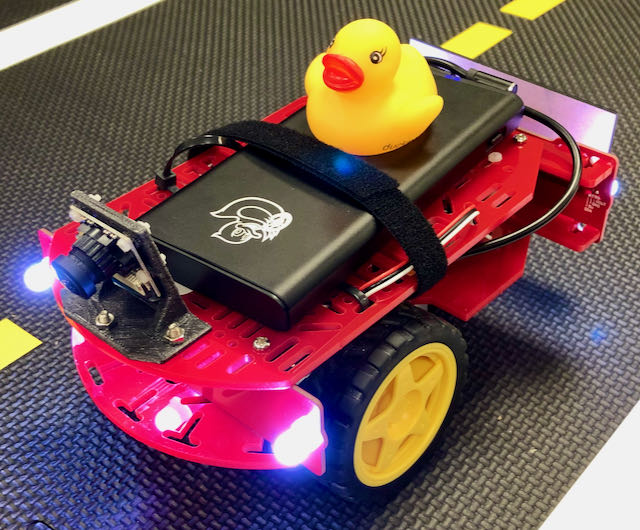}
  \caption{Duckietown DB18 robot.}
  \label{fig:DB18}
\end{figure}

\subsection{Hardware and Software Platform}

In our experiments the goal is to demonstrate that the $\beta$-VAE based OOD detector (presented briefly in Section~\ref{sec:ood}) could be run on a CPS platform with limited hardware capabilities and resource scheduling constraints. We selected the Duckietown DB18~\cite{db18}, which uses a Rasberry Pi 3B+~\cite{db18} as its primary computation unit. The Rasberry Pi 3B+ is equipped with a Broadcom BCM2837B0 quad-core A53 (ARMv8) 64-bit CPU operating at 1.4GHz. In our experiment we only use one sensor on the DB18, the front facing camera with a field-of-view of 160 degrees. This camera provides data to both the autonomous driving and OOD detection subsystems. In addition, we are only concerned with controlling the outputs of two actuators: the left and right wheel motors. Figure~\ref{fig:DB18} shows a depiction of the physical robot we used for these experiments.

\begin{figure}
  \centering
  \includegraphics[width=\linewidth]{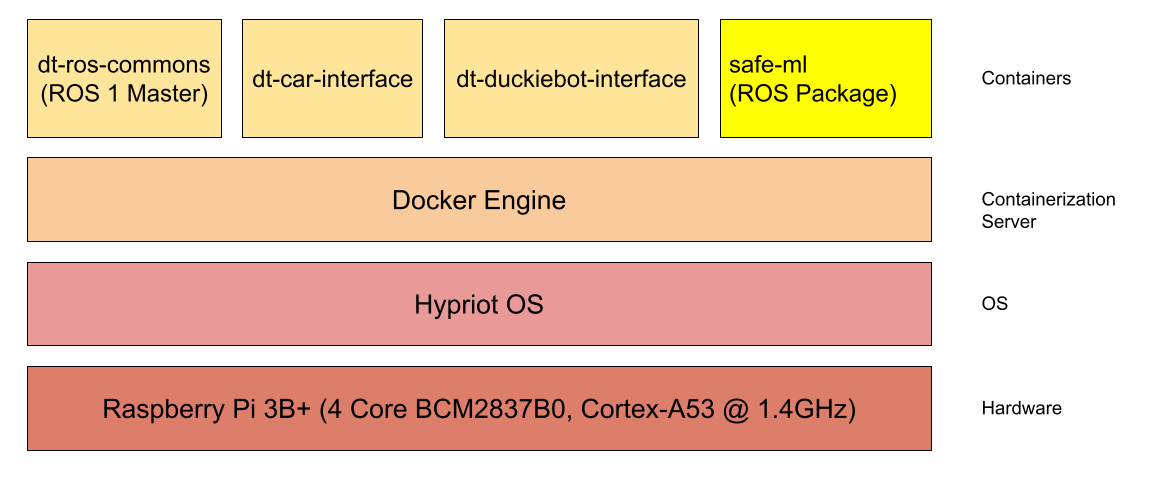}
  \caption{Depiction of the Duckietown software stack and integration with the ROS packages developed for this research.}
  \label{fig:software_abs}
\end{figure}

The Duckietown platform also provides a software framework that takes advantage of ROS and Docker to allow for fast, modular development. Figure~\ref{fig:software_abs} provides a depiction of the Duckietown software stack. The Rasberry Pi is running Hypriot OS bare-metal~\cite{hypriot}, which in turn runs the Docker engine. Docker allows for containerization of applications and helps to manage dependencies as well as ensure that our architecture remains modular~\cite{gonzalez2017modular}. ROS controls and manages communication between nodes and services that provide abstraction for the hardware layer. Duckietown has also provided built-in ROS nodes and services to interface with the vehicle hardware to help maintain modularity and enforce separation of concerns.

\subsection{System Architecture}

\begin{figure}[h]
  \centering
  \includegraphics[width=0.9\linewidth]{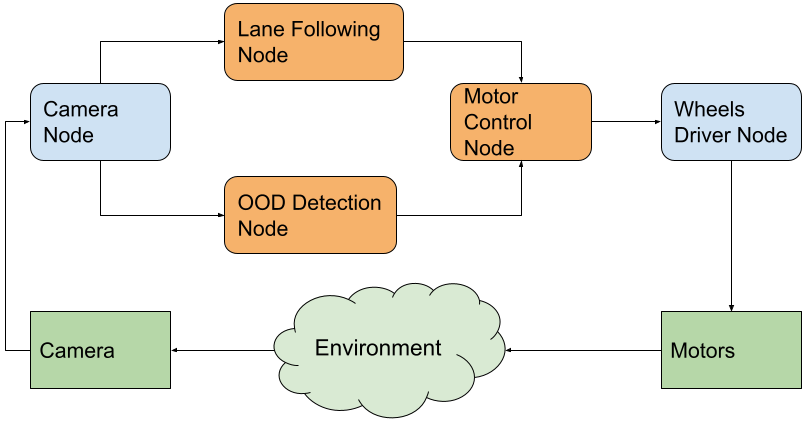}
  \caption{Block diagram of OOD detection architecture utilizing existing Duckietown framework.}
  \label{fig:architecture}
\end{figure}

Figure~\ref{fig:architecture} shows a block diagram of our system architecture. Each block corresponds to a ROS node. The color coding is as follows: blue nodes are provided to us by the Duckietown framework, orange nodes were implemented by us, and the green blocks correspond to physical sensors and actuators. The camera node provides a ROS interface to the camera hardware which captures the state of the environment in front of the Duckiebot at a rate of $30 \mbox{fps}$.  The compressed image data as well as the time stamp of when it was collected are passed to other nodes in the system so that we are able to make measurements on the throughput and latency of our architecture in addition to demonstrating functionality. 

\vspace{-3mm}
The lane following node subscribes to the camera node and generates a suggested control action to steer the Duckiebot while traveling in the physical environment (navigation system). Due to resource constraints on the DB18, we implemented this node using classical computer vision techniques~\cite{lanenavigation}. Images from the Duckiebot's camera are first resized into images of size 128x96 pixels from 640x480 pixels. We also isolated and used only the lower half of the image to reduce the amount of computation required. A mask is then applied to filter the white lanes out from the image. Next, we use the Canny edge detection algorithm to extract all edges from the white areas and perform a Hough Transform to identify the line segments from the image. Thereafter, the line segments are combined into their left and right lane lines based on their slopes. The detected angle from the lane lines are then denoised by an averaging filter and fed to the motor control node to control the Duckiebot's steering.

The OOD detection node also subscribes to the camera node. Its goal is to analyze incoming images and identify any that are outside of the training dataset (as described in Section~\ref{sec:ood}). Note, although the navigation system in our implementation does not rely on this training dataset, nevertheless we believe that the implemented framework is sufficient to demonstrate and evaluate the effectiveness of OOD detection algorithms. These algorithms are completely independent of the navigation systems, except that they share the training dataset with the navigation systems when those systems are implemented using ML techniques.  

The motor control node serves to generate the desired velocities for each wheel. It takes the desired steering angle generated by the lane following node and applies a PID controller to shape the desired steering response.  It then calculates desired wheel velocities based on the output of this PID controller. In addition, it is subscribed to an emergency stop topic provided by the OOD detection node. When an emergency stop message is received upon detection of an OOD image, a callback is executed which will immediately set all wheel velocities to zero until the robot comes to a halt. The motor control node passes the calculated wheel velocities to the wheels driver node provided by the Duckietown framework, which interfaces with the physical motors.

\section{Experiment Design}
\label{sec:experiment}

We implemented the system architecture described in Section~\ref{sec:framework} and tested the performance of this system in an emergency stop scenario. Our goal was to evaluate the run-time performance of the OOD detection algorithm, both in terms of its accuracy in detecting OOD images as well as the timing delays in the system. This section describes the specifics of the experiment setup.

\subsection{OOD Detector}

We conducted OOD detection in $\beta$-VAE latent space, as proposed in~\cite{ramakrishna2020efficient} and briefly summarized in Section~\ref{sec:ood}. The $\beta$-VAE encoder has four CNN layers of 32/64/128 x (5x5) filters with exponential linear unit (ELU) activation and batchnorm layers, two 2x2 max-pooling layers, and one fully connected layer of size 1568 with ELU activation. The decoder is architecturally symmetric. 

\begin{figure}
  \centering
  \includegraphics[width=\linewidth]{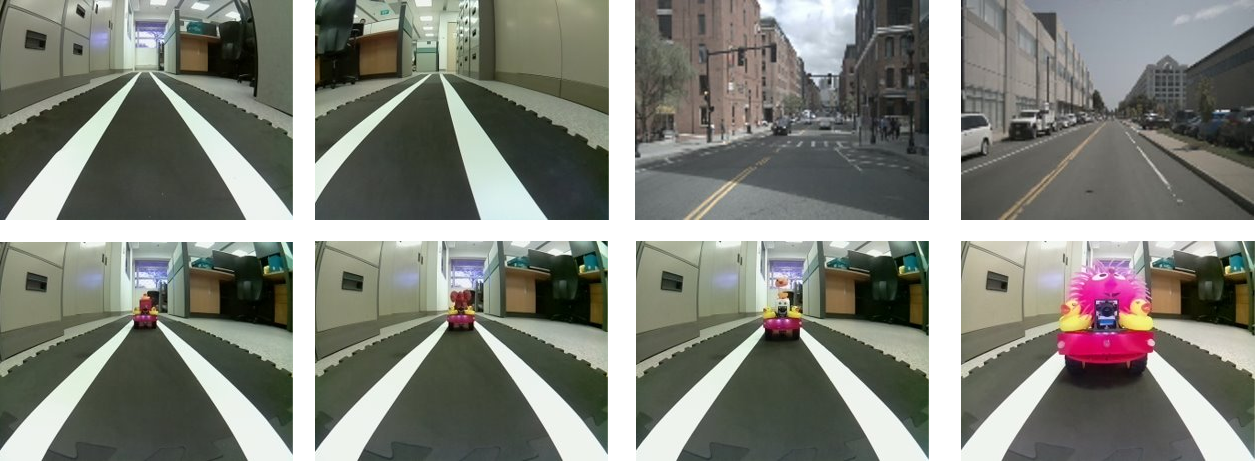}
  \caption{Sample in-distribution and OOD images from the experiment. The top row is in-distribution images from the Duckiebot test bed (top left) and nuScenes (top right). The bottom row shows OOD images of obstacles blocking the Duckiebot's path at distances of 75, 45, 30 and 15 cm from the left to right.}
  \label{fig:testbed}  
\end{figure}

To avoid overfitting and reduce the data collection effort, we collected in-distribution data from two sources to train a $\beta$-VAE model. First, we used the real-world driving scenes from nuScenes mini~\cite{caesar2020nuscenes} as the main, in-distribution training data of $2350$ images.  Following the hyperparameters proposed in~\cite{ramakrishna2020efficient}, we set the dimensions of the latent space to $30$ and the $\beta$ value to $1.4$, then used the Adam optimizer in PyTorch to train the $\beta$-VAE model for $100$ epochs with a learning rate of $10^{-5}$. Subsequently, we used the Rosbag tool to capture a small, in-distribution set of $1240$ images from our Duckiebot test bed, shuffled the sequence, and divided the set into a fine-tuning set and an equally sized calibration set. The model trained with nuScenes data was further trained with the fine-tuning set for $10$ epochs with the same learning rate. The top row of Figure~\ref{fig:testbed} shows examples of in-distribution images.

We define in-distribution data as images in which the lane in front of the autonomous bot is clear and not blocked. If there is an obstacle in the lane in front of the Duckiebot, we define the image as OOD. The obstacles used for testing are shown in the bottom row of Figure~\ref{fig:testbed}. From left to right, the Duckiebot was driving toward an obstacle placed in its driving lane.

\subsection{Navigation and Emergency Braking System}

As described in Section~\ref{sec:framework}, due to resource constraints, we used an OpenCV-based lane following algorithm to navigate the Duckiebot down the test track. The cycle time for this controller was set at $5 Hz$ which was empirically determined to be a good compromise between accuracy in lane-line detection and resource utilization.
\begin{figure}
  \centering
  \includegraphics[width=0.8\linewidth]{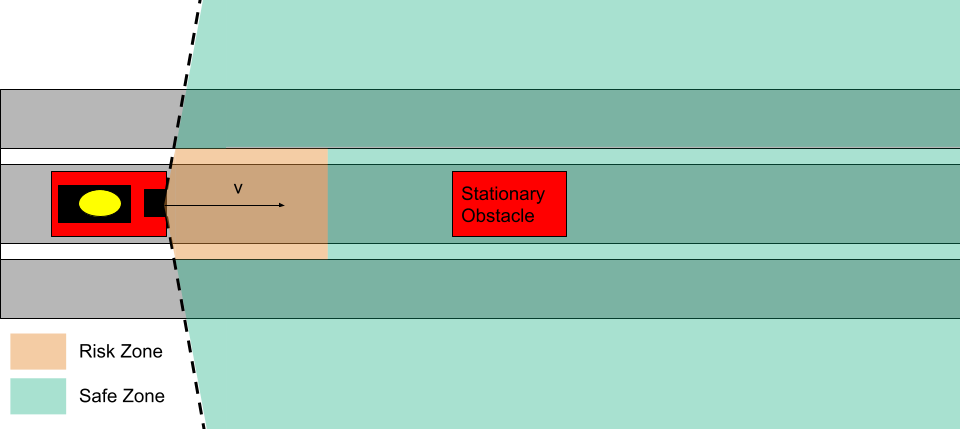}
  \caption{Emergency braking experiment setup. The Duckiebot moves at constant velocity until a stationary obstacle enters its risk zone, at which point it must stop before colliding with the obstacle.}
  \label{fig:setup}
\end{figure}
The camera node sent compressed images to the OOD detector and lane follower using ROS topics. The nominal frame rate was measured at $30 \mbox{fps}$. Camera calibration was performed according to the Duckietown specification before conducting the experiment.

Figure~\ref{fig:setup} depicts the overall experiment setup. We partition the area in front of the Duckiebot running OOD detection into two zones: the \emph{safe zone} and the \emph{risk zone}.  The safe zone is a region sufficiently far from the bot where even if an obstacle is present, its physical distance is sufficient that it does not pose an immediate threat to bot safety. The risk zone is defined as a region adjacent to the bot in which the presence of an obstacle poses an immediate threat to safety and the bot must therefore be stopped.

The Duckiebot is placed stationary at a fixed location on its track and an obstacle is placed $70cm$ away in the center of the lane. The Duckiebot is then driven forward using it's autonomous lane following algorithm. Simultaneously, the OOD detector is generating an OOD score. When the OOD score surpasses a fixed \emph{OOD detection threshold} (defined experimentally as greater than the $80^{th}$ percentile of the scores of in-distribution images), the OOD detector will send an emergency stop signal to the motor control and the Duckiebot should come to a stop. It is important to note that the image processed by the OOD detector is always the latest image available from the camera;  intermediate images captured while the OOD detector is still processing an old image are dropped. This means that the OOD detector does not process every image captured by the camera due to resource constraints.

The distance between the final stopping position of the bot and the obstacle is then recorded as the \emph{stopping distance}. This distance is measured from the leading edge of the Duckiebot to the leading edge of the obstacle such that a distance of $0cm$ indicates a collision. The time from when the obstacle entered the Duckiebot's risk zone to the emergency stop was also measured (\emph{end-to-end stopping time}). To get an accurate measurement without affecting the performance of the moving Duckiebot, we placed obstacles on top of a second Duckiebot recording a video of the oncoming vehicle. By analyzing this video we were able to determine the time between the obstacle entering the mobile bot's risk zone and the emergency stop with a resolution of $50ms$, limited by the frame capture rate of the stationary bot's camera. In addition to stopping distance and end-to-end stopping time, we also measured the duration it took to capture an image and send it to the OOD detection algorithm (defined as \emph{image capture to detection start time}), the running time of the OOD detection algorithm itself (defined as \emph{OOD detector execution time}), and the duration between an OOD image being detected and the motor velocity being set to zero (defined as \emph{detection result to motor stop time}). Here \emph{image capture time} refers to the timestamps when an image was captured by the Duckiebot's camera, but before it was processed. Using timestamps for image capture time and the total distance the Duckiebot traveled in each test run, we were able to generate a velocity estimate and use this value to estimate the distance traveled when each image capture event occurred.

We repeated $40$ runs of this experiment with $10$ runs each using $4$ different types of obstacles. Across these $40$ experiment runs the OOD detection algorithm ran $288$ times, with a minimum of $2$ executions and a maximum of $11$ executions for each run.

\section{Results}
\label{sec:results}

In this section, results of our experiments are discussed, as well as key inferences that can be drawn from our data are presented.

\subsection{Overall Performance of OOD Detection}

Using the OOD detection threshold set at $80\%$ of the training data set's OOD scores, we found that out of $40$ test runs we had $5$ collisions with the stationary obstacle resulting in a successful emergency stopping rate of $87.5\%$. We found that the median stopping distance value was $14.5cm$ with a $95\%$ confidence interval of $[9.0cm, 20.0cm]$. This result indicates that the $\beta$-VAE detector running on the Duckiebot was, in the general case, able to successfully stop the bot in response to an OOD condition on the road ahead; collision cases are further discussed in the subsequent section.

\begin{figure}[h]
  \centering
  \includegraphics[width=\linewidth]{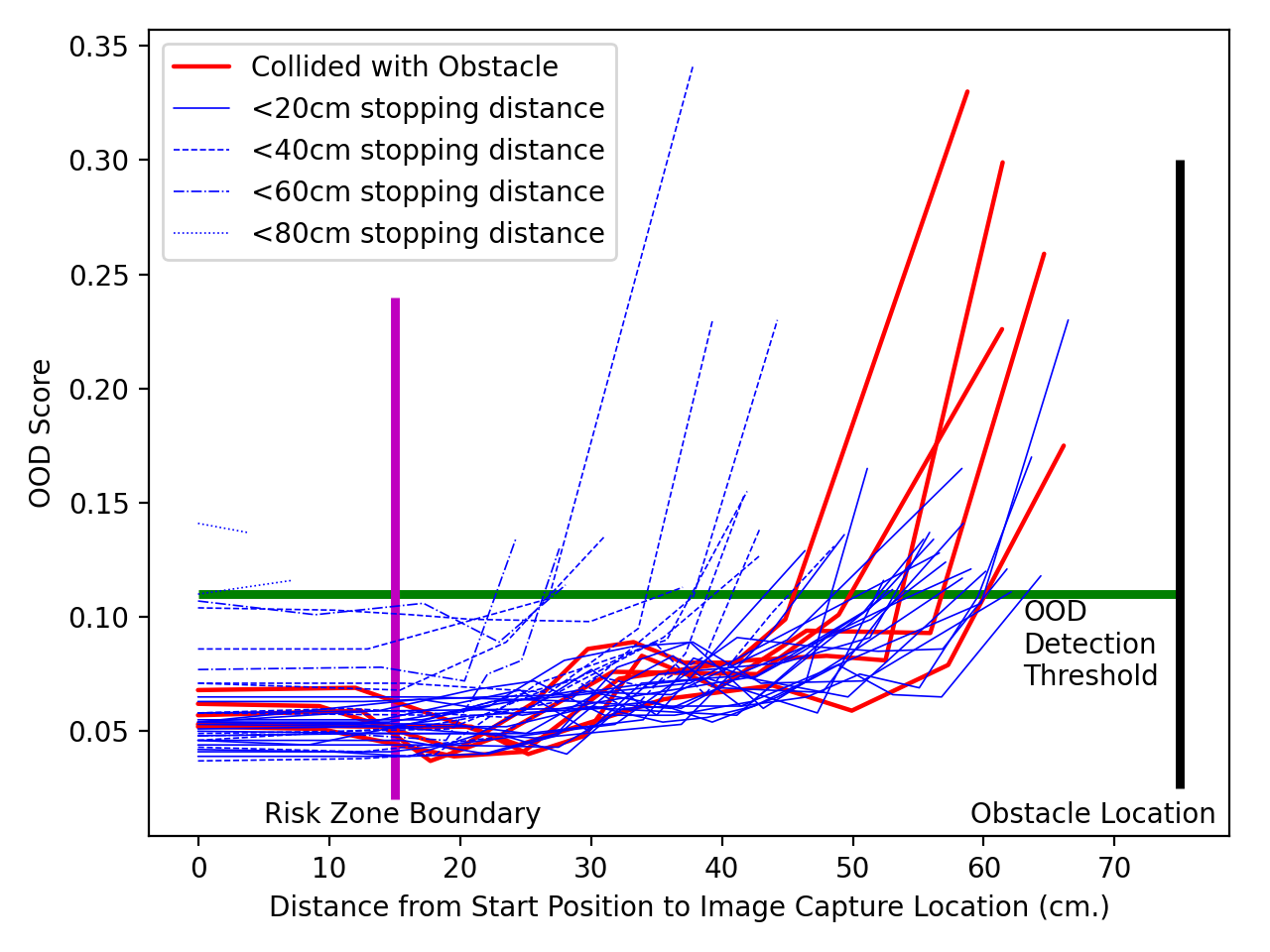}
  \caption{Distribution of OOD scores as a function of distance from the starting position. The scores are based on images that are captured at these distances. Each line denotes all of the OOD scores during a specific test run. Line styles are used to classify the various runs based on the achieved stopping distance from the obstacle.}
  \label{fig:DTvsOOD}
\vspace{-1mm}
\end{figure}

Figure~\ref{fig:DTvsOOD} shows the OOD scores for images gathered at various distances from the starting position. This distance was not measured directly, but was calculated based on the Duckiebot's estimated velocity as discussed in Section~\ref{sec:experiment}. Each line corresponds to one run of the experiment. A red line indicates that the run ended in a collision with an obstacle. A blue line indicates that the Duckiebot stopped successfully before the obstacle.  Line style variation is used to indicate how far from the obstacle the Duckiebot stopped. The green line represents the detection threshold for OOD score. The black line represents the location of the obstacle and the magenta line shows the risk zone boundary. It is important to note that the distances shown in this figure are based on the image capture time. The OOD score will not be available at the distances shown; they will only be available at a later time (closer to the obstacle) when the OOD detector actually completes execution.

From Figure~\ref{fig:DTvsOOD} we can observe that every test run had at least one image for which the OOD score was above the threshold; even for the runs that resulted in a collision. Unfortunately, in these collision cases, the OOD outcome was only available after the Duckiebot hit the obstacle. One can also observe that the image which led to a high OOD score was captured earlier in the case of some runs with collision when compared to some other runs without collision. This indicates the presence of high variabilities in the execution times of the OOD detection algorithm. Finally, from the figure we can observe that in $2$ runs the Duckiebot stopped between $80cm$ and $60cm$ from the obstacle (outside the risk zone), in $13$ runs it stopped between $60cm$ and $20cm$ from the obstacle (inside the risk zone), and in $20$ runs it stopped within $20cm$ of the obstacle. 

\vspace{-4mm}
\subsection{Analysis of Collision Cases}
\label{sec:results_failures}

Based on the data in Figure~\ref{fig:DTvsOOD} we observe that in each of the $40$ test runs, there is at least one input image with a high OOD score captured before the bot reaches the obstacle. We wanted to understand why a collision still occurred in $5$ of those runs. One possible source of error is that our system was using ROS running on a variant of Linux \textit{without} any real-time kernel patch. Without hard real-time guarantees from the scheduler, execution time for OOD detection related tasks could be variable and lead to delayed detection and control actions from the Duckiebot.

\begin{figure}[htbp]
  \centering
  \includegraphics[width=\linewidth]{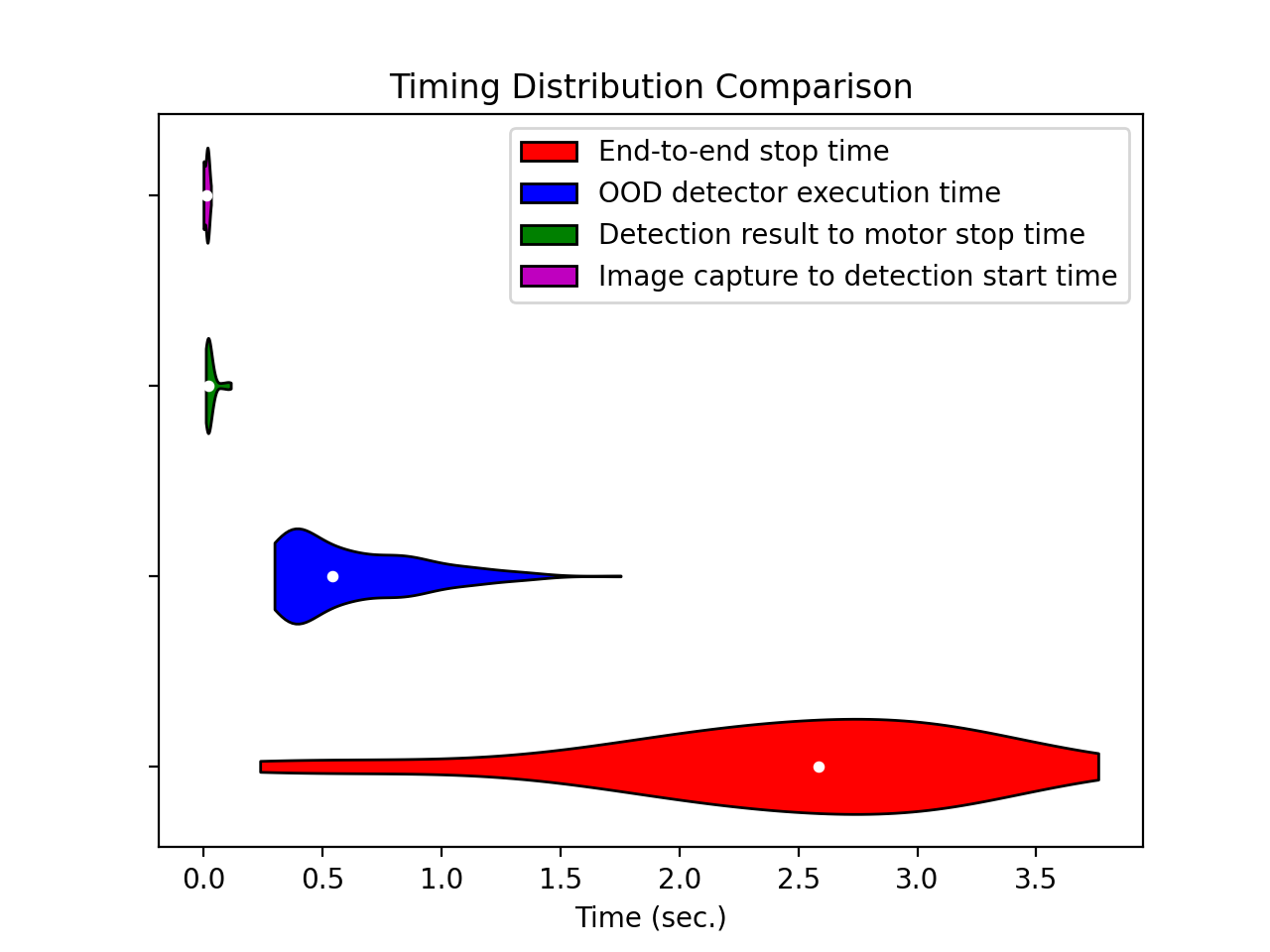}
  \caption{Violin plot of distribution of sub-task execution times for all the test runs.}
  \label{fig:dist_fa}
\end{figure}

Figure~\ref{fig:dist_fa} shows the distribution of execution times for various tasks in our implementation and compares them to the end-to-end stopping time of the Duckiebot, that is the duration from when the obstacle enters the risk zone to when the bot comes to a physical stop. The tasks we observed were the time for the OOD detector to execute, the time from a positive OOD detection result to the motor stop command being issued, and the time from an image being captured by the camera to its ingestion by the OOD detection algorithm. It can be observed from the figure that a majority of the delay was spent waiting for the OOD detector to complete execution, and that OOD detector execution times have a long tail distribution skewed towards longer detection times. In comparison, the time spent for passing images and motor commands via ROS topics is inconsequential.

\begin{table}[htbp]
  \caption{OOD detector execution times preceding collisions}
  \label{tab:fails}
  \begin{tabular}{ccl}
    \toprule
    Failure&Last OOD exec. time&Penultimate OOD exec. time\\
    \midrule
    Case 1&$1.328 \mbox{sec}$.&$1.202 \mbox{sec}$.\\
    Case 2&$0.846 \mbox{sec}$.&$0.721 \mbox{sec}$.\\
    Case 3&$0.830 \mbox{sec}$.&$0.885 \mbox{sec}$.\\
    Case 4&$0.621 \mbox{sec}$.&$0.664 \mbox{sec}$.\\
    Case 5&$1.073 \mbox{sec}$.&$0.747 \mbox{sec}$.\\
  \bottomrule
\end{tabular}
\vspace{-2mm}
\end{table}

In Table~\ref{tab:fails} it can be observed that our $5$ failure cases all had at least two preceding OOD detector execution times longer than the overall median execution time of $542 \mbox{sec}$. This underlines the importance of using hard real-time OSs in safety critical applications. Given the constraints of the system we were using, we conjecture that a decreased OOD detection threshold or reduced bot speed could help mitigate for this limitation. However, these adjustments come with some side-effects (e.g., increased false positives and reduced system performance), and these trade-offs must be carefully studied.

\subsection{Impact of OOD Detection Threshold on Performance}

\begin{figure}[h]
\vspace{-1mm}
  \centering
  \includegraphics[width=0.8\linewidth]{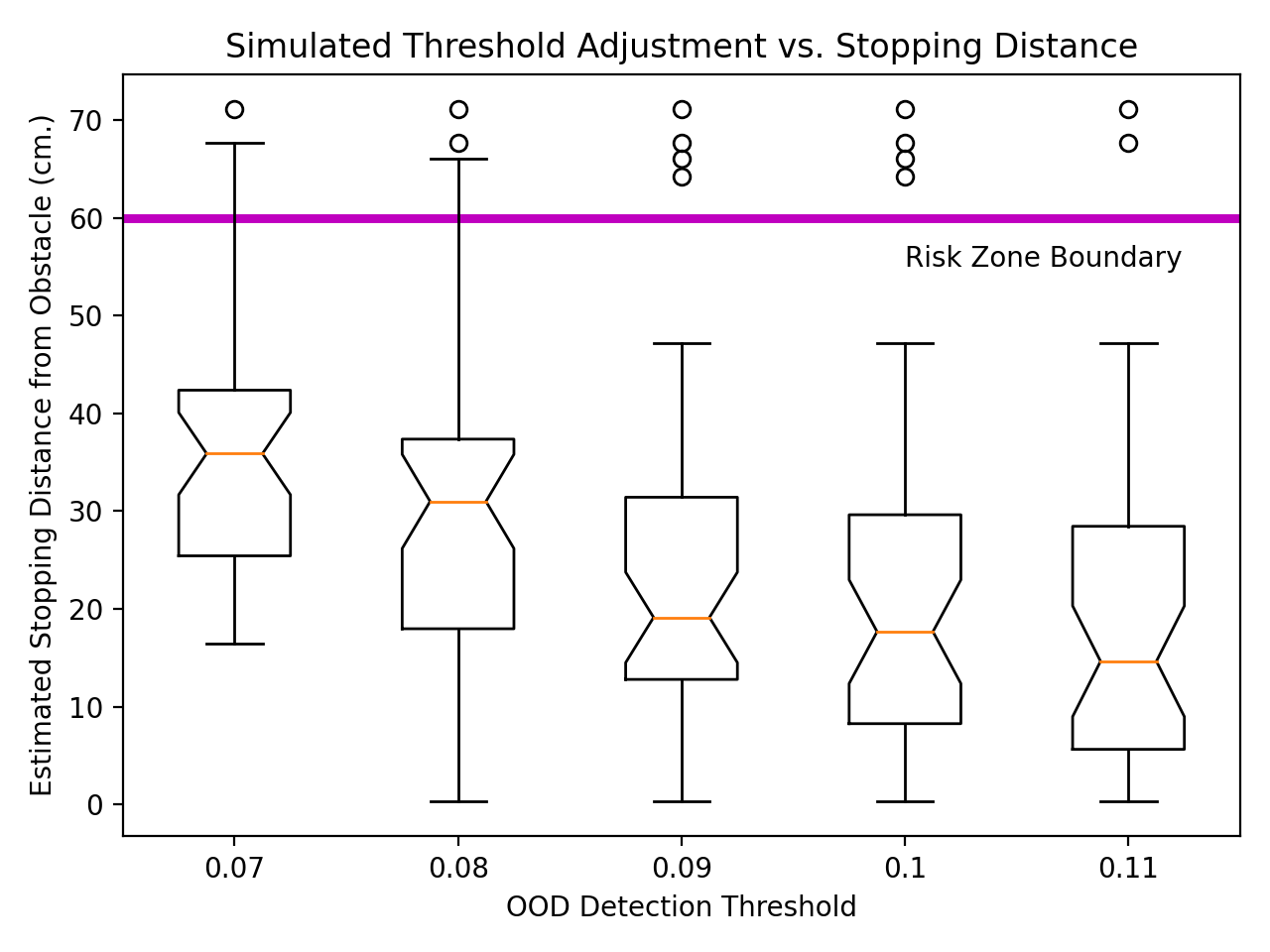}\vspace{-2mm}
  \caption{Boxplots with confidence interval showing the distribution of projected stopping distances for varying OOD detection thresholds. The middle-quartile line marks the estimated median value.}
  \label{fig:TH_SIM}
\vspace{-3mm}
\end{figure}

In this section, we explore how adjusting the OOD detection threshold impacts the stopping distance of the Duckiebot from the obstacle. Using the OOD scores and stopping distances gathered during the $40$ test runs, we simulated the impact on this stopping distance for a variety of threshold values using the velocity estimate described in Section~\ref{sec:experiment}. Figure~\ref{fig:TH_SIM} shows the distribution of simulated stopping distances for several thresholds. It can be observed that as the threshold is lowered, the Duckiebot is projected to stop on average further from the obstacle as expected. While this decreases the projected number of collisions, it comes with a trade-off: as the threshold is lowered, more images outside the risk zone are detected as OOD (above the risk zone line shown in the figure). This would result in the Duckiebot stopping even when there is no risk of collision. Conceptually this makes sense, as an OOD detection threshold of $0$ would result in all images being classified as OOD and although the Duckiebot would never collide with an obstacle, it would also never move. Likewise, increasing the OOD detection threshold reduces the likelihood of these false positives, but brings the projected stopping distance closer to the obstacle, where the timing variations described in Section~\ref{sec:results_failures} have a greater chance to cause a collision.

\vspace{-2mm}
\vspace{-3mm}

\section{Conclusions}

We successfully demonstrated that the $\beta$-VAE OOD detection algorithm could run on an embedded platform and provides a safety check on the control of an autonomous robot. We also showed that performance is dependent on real-time performance of the embedded system, particularly the OOD detector execution time. Lastly, we showed that there is a trade-off involved in choosing an OOD detection threshold; a smaller threshold value increases the average stopping distance from an obstacle, but leads to an increase in false positives.

This work also generates new questions that we hope to investigate in the future. The system architecture demonstrated in this paper was not utilizing a real-time OS and did not take advantage of technologies such as GPUs or TPUs, which are now becoming common on embedded systems. There is still much work that can be done to optimize process scheduling and resource utilization while maintaining the goal of using low-cost, off-the-shelf hardware and open-source software. Understanding what quality of service can be provided by a system with these constraints and whether it suffices for reliable operations of OOD detection algorithms is an ongoing research theme.

From the OOD detection perspective, we would like to run additional OOD detection algorithms on the same architecture and compare performance in terms of accuracy and computational efficiency. We would also like to develop a more comprehensive set of test scenarios to serve as a benchmark for OOD detection on embedded systems. These should include dynamic as well as static obstacles, operation in various environments and lighting conditions, and OOD scenarios that occur while the robot is performing more complex tasks like navigating corners, intersections, or merging with other traffic.

Demonstrating OOD detection on the Duckietown platform opens the door for more embedded applications of OOD detectors. This will serve to better evaluate their usefulness as a tool to enhance the safety of ML systems deployed as part of critical CPS.

\vspace{-1mm}
\begin{acks}
This research was funded in part by MoE, Singapore, Tier-2 grant number MOE2019-T2-2-040.
\end{acks}
\vspace{-1mm}

\bibliographystyle{ACM-Reference-Format}
\bibliography{sample-base}


\begin{thebibliography}{20}


\ifx \showCODEN    \undefined \def \showCODEN     #1{\unskip}     \fi
\ifx \showDOI      \undefined \def \showDOI       #1{#1}\fi
\ifx \showISBNx    \undefined \def \showISBNx     #1{\unskip}     \fi
\ifx \showISBNxiii \undefined \def \showISBNxiii  #1{\unskip}     \fi
\ifx \showISSN     \undefined \def \showISSN      #1{\unskip}     \fi
\ifx \showLCCN     \undefined \def \showLCCN      #1{\unskip}     \fi
\ifx \shownote     \undefined \def \shownote      #1{#1}          \fi
\ifx \showarticletitle \undefined \def \showarticletitle #1{#1}   \fi
\ifx \showURL      \undefined \def \showURL       {\relax}        \fi
\providecommand\bibfield[2]{#2}
\providecommand\bibinfo[2]{#2}
\providecommand\natexlab[1]{#1}
\providecommand\showeprint[2][]{arXiv:#2}

\bibitem[\protect\citeauthoryear{{Almási}, {Moni}, and
  {Gyires-Tóth}}{{Almási} et~al\mbox{.}}{2020}]%
        {9207497}
\bibfield{author}{\bibinfo{person}{P. {Almási}}, \bibinfo{person}{R. {Moni}},
  {and} \bibinfo{person}{B. {Gyires-Tóth}}.} \bibinfo{year}{2020}\natexlab{}.
\newblock \showarticletitle{Robust Reinforcement Learning-based Autonomous
  Driving Agent for Simulation and Real World}. In
  \bibinfo{booktitle}{\emph{2020 International Joint Conference on Neural
  Networks (IJCNN)}}. \bibinfo{pages}{1--8}.
\newblock
\urldef\tempurl%
\url{https://doi.org/10.1109/IJCNN48605.2020.9207497}
\showDOI{\tempurl}


\bibitem[\protect\citeauthoryear{Bulusu, Kailkhura, Li, Varshney, and
  Song}{Bulusu et~al\mbox{.}}{2020}]%
        {bulusu2020anomalous}
\bibfield{author}{\bibinfo{person}{Saikiran Bulusu}, \bibinfo{person}{Bhavya
  Kailkhura}, \bibinfo{person}{Bo Li}, \bibinfo{person}{Pramod~K Varshney},
  {and} \bibinfo{person}{Dawn Song}.} \bibinfo{year}{2020}\natexlab{}.
\newblock \showarticletitle{Anomalous Instance Detection in Deep Learning: A
  Survey}.
\newblock \bibinfo{journal}{\emph{arXiv preprint arXiv:2003.06979}}
  (\bibinfo{year}{2020}).
\newblock


\bibitem[\protect\citeauthoryear{Caesar, Bankiti, Lang, Vora, Liong, Xu,
  Krishnan, Pan, Baldan, and Beijbom}{Caesar et~al\mbox{.}}{2020}]%
        {caesar2020nuscenes}
\bibfield{author}{\bibinfo{person}{Holger Caesar}, \bibinfo{person}{Varun
  Bankiti}, \bibinfo{person}{Alex~H Lang}, \bibinfo{person}{Sourabh Vora},
  \bibinfo{person}{Venice~Erin Liong}, \bibinfo{person}{Qiang Xu},
  \bibinfo{person}{Anush Krishnan}, \bibinfo{person}{Yu Pan},
  \bibinfo{person}{Giancarlo Baldan}, {and} \bibinfo{person}{Oscar Beijbom}.}
  \bibinfo{year}{2020}\natexlab{}.
\newblock \showarticletitle{nuscenes: A multimodal dataset for autonomous
  driving}. In \bibinfo{booktitle}{\emph{Proceedings of the IEEE/CVF conference
  on computer vision and pattern recognition}}. \bibinfo{pages}{11621--11631}.
\newblock


\bibitem[\protect\citeauthoryear{Foundation}{Foundation}{[n.d.]a}]%
        {db18}
\bibfield{author}{\bibinfo{person}{The~Duckietown Foundation}.}
  \bibinfo{year}{[n.d.]}\natexlab{a}.
\newblock \bibinfo{booktitle}{\emph{The Duckiebot Manual - Unit C-4 -
  Assembling the Duckiebot DB18}}.
\newblock
\urldef\tempurl%
\url{https://docs.duckietown.org/daffy/opmanual_duckiebot/out/assembling_duckiebot_db18.html}
\showURL{%
\tempurl}


\bibitem[\protect\citeauthoryear{Foundation}{Foundation}{[n.d.]b}]%
        {hypriot}
\bibfield{author}{\bibinfo{person}{The~Duckietown Foundation}.}
  \bibinfo{year}{[n.d.]}\natexlab{b}.
\newblock \bibinfo{booktitle}{\emph{The Duckiebot Manual - Unit C-8 - Duckiebot
  SD Card Initialization}}.
\newblock
\urldef\tempurl%
\url{https://docs.duckietown.org/daffy/opmanual_duckiebot/out/setup_duckiebot.html}
\showURL{%
\tempurl}


\bibitem[\protect\citeauthoryear{Gonz{\'a}lez-Nalda, Etxeberria-Agiriano,
  Calvo, and Otero}{Gonz{\'a}lez-Nalda et~al\mbox{.}}{2017}]%
        {gonzalez2017modular}
\bibfield{author}{\bibinfo{person}{Pablo Gonz{\'a}lez-Nalda},
  \bibinfo{person}{Ismael Etxeberria-Agiriano}, \bibinfo{person}{Isidro Calvo},
  {and} \bibinfo{person}{Mari~Carmen Otero}.} \bibinfo{year}{2017}\natexlab{}.
\newblock \showarticletitle{A modular CPS architecture design based on ROS and
  Docker}.
\newblock \bibinfo{journal}{\emph{International Journal on Interactive Design
  and Manufacturing (IJIDeM)}} \bibinfo{volume}{11}, \bibinfo{number}{4}
  (\bibinfo{year}{2017}), \bibinfo{pages}{949--955}.
\newblock


\bibitem[\protect\citeauthoryear{Gu and Easwaran}{Gu and Easwaran}{2019}]%
        {gu2019towards}
\bibfield{author}{\bibinfo{person}{Xiaozhe Gu} {and} \bibinfo{person}{Arvind
  Easwaran}.} \bibinfo{year}{2019}\natexlab{}.
\newblock \showarticletitle{Towards safe machine learning for cps: infer
  uncertainty from training data}. In \bibinfo{booktitle}{\emph{Proceedings of
  the 10th ACM/IEEE International Conference on Cyber-Physical Systems}}.
  \bibinfo{pages}{249--258}.
\newblock


\bibitem[\protect\citeauthoryear{Higgins, Matthey, Pal, Burgess, Glorot,
  Botvinick, Mohamed, and Lerchner}{Higgins et~al\mbox{.}}{2016}]%
        {higgins2016beta}
\bibfield{author}{\bibinfo{person}{Irina Higgins}, \bibinfo{person}{Loic
  Matthey}, \bibinfo{person}{Arka Pal}, \bibinfo{person}{Christopher Burgess},
  \bibinfo{person}{Xavier Glorot}, \bibinfo{person}{Matthew Botvinick},
  \bibinfo{person}{Shakir Mohamed}, {and} \bibinfo{person}{Alexander
  Lerchner}.} \bibinfo{year}{2016}\natexlab{}.
\newblock \showarticletitle{beta-vae: Learning basic visual concepts with a
  constrained variational framework}.
\newblock  (\bibinfo{year}{2016}).
\newblock


\bibitem[\protect\citeauthoryear{{Kalapos}, {Gór}, {Moni}, and
  {Harmati}}{{Kalapos} et~al\mbox{.}}{2020}]%
        {9263751}
\bibfield{author}{\bibinfo{person}{A. {Kalapos}}, \bibinfo{person}{C. {Gór}},
  \bibinfo{person}{R. {Moni}}, {and} \bibinfo{person}{I. {Harmati}}.}
  \bibinfo{year}{2020}\natexlab{}.
\newblock \showarticletitle{Sim-to-real reinforcement learning applied to
  end-to-end vehicle control}. In \bibinfo{booktitle}{\emph{2020 23rd
  International Symposium on Measurement and Control in Robotics (ISMCR)}}.
  \bibinfo{pages}{1--6}.
\newblock
\urldef\tempurl%
\url{https://doi.org/10.1109/ISMCR51255.2020.9263751}
\showDOI{\tempurl}


\bibitem[\protect\citeauthoryear{Kato, Takeuchi, Ishiguro, Ninomiya, Takeda,
  and Hamada}{Kato et~al\mbox{.}}{2015}]%
        {kato2015open}
\bibfield{author}{\bibinfo{person}{Shinpei Kato}, \bibinfo{person}{Eijiro
  Takeuchi}, \bibinfo{person}{Yoshio Ishiguro}, \bibinfo{person}{Yoshiki
  Ninomiya}, \bibinfo{person}{Kazuya Takeda}, {and} \bibinfo{person}{Tsuyoshi
  Hamada}.} \bibinfo{year}{2015}\natexlab{}.
\newblock \showarticletitle{An open approach to autonomous vehicles}.
\newblock \bibinfo{journal}{\emph{IEEE Micro}} \bibinfo{volume}{35},
  \bibinfo{number}{6} (\bibinfo{year}{2015}), \bibinfo{pages}{60--68}.
\newblock


\bibitem[\protect\citeauthoryear{Kingma and Welling}{Kingma and
  Welling}{2013}]%
        {kingma2013auto}
\bibfield{author}{\bibinfo{person}{Diederik~P Kingma} {and}
  \bibinfo{person}{Max Welling}.} \bibinfo{year}{2013}\natexlab{}.
\newblock \showarticletitle{Auto-encoding variational bayes}.
\newblock \bibinfo{journal}{\emph{arXiv preprint arXiv:1312.6114}}
  (\bibinfo{year}{2013}).
\newblock


\bibitem[\protect\citeauthoryear{Koubaa}{Koubaa}{2017}]%
        {ros_vol2}
\bibfield{author}{\bibinfo{person}{Anis Koubaa}.}
  \bibinfo{year}{2017}\natexlab{}.
\newblock \bibinfo{booktitle}{\emph{Robot Operating System (ROS): The Complete
  Reference (Volume 2)} (\bibinfo{edition}{1st} ed.)}.
\newblock \bibinfo{publisher}{Springer Publishing Company, Incorporated}.
\newblock
\showISBNx{331954926X}


\bibitem[\protect\citeauthoryear{Merkel}{Merkel}{2014}]%
        {docker}
\bibfield{author}{\bibinfo{person}{Dirk Merkel}.}
  \bibinfo{year}{2014}\natexlab{}.
\newblock \showarticletitle{Docker: Lightweight Linux Containers for Consistent
  Development and Deployment}.
\newblock \bibinfo{journal}{\emph{Linux J.}} \bibinfo{volume}{2014},
  \bibinfo{number}{239}, Article \bibinfo{articleno}{2} (\bibinfo{date}{March}
  \bibinfo{year}{2014}).
\newblock
\showISSN{1075-3583}


\bibitem[\protect\citeauthoryear{{Paull}, {Tani}, {Ahn}, {Alonso-Mora},
  {Carlone}, {Cap}, {Chen}, {Choi}, {Dusek}, {Fang}, {Hoehener}, {Liu},
  {Novitzky}, {Okuyama}, {Pazis}, {Rosman}, {Varricchio}, {Wang}, {Yershov},
  {Zhao}, {Benjamin}, {Carr}, {Zuber}, {Karaman}, {Frazzoli}, {Del Vecchio},
  {Rus}, {How}, {Leonard}, and {Censi}}{{Paull} et~al\mbox{.}}{2017}]%
        {duckietown}
\bibfield{author}{\bibinfo{person}{L. {Paull}}, \bibinfo{person}{J. {Tani}},
  \bibinfo{person}{H. {Ahn}}, \bibinfo{person}{J. {Alonso-Mora}},
  \bibinfo{person}{L. {Carlone}}, \bibinfo{person}{M. {Cap}},
  \bibinfo{person}{Y.~F. {Chen}}, \bibinfo{person}{C. {Choi}},
  \bibinfo{person}{J. {Dusek}}, \bibinfo{person}{Y. {Fang}},
  \bibinfo{person}{D. {Hoehener}}, \bibinfo{person}{S. {Liu}},
  \bibinfo{person}{M. {Novitzky}}, \bibinfo{person}{I.~F. {Okuyama}},
  \bibinfo{person}{J. {Pazis}}, \bibinfo{person}{G. {Rosman}},
  \bibinfo{person}{V. {Varricchio}}, \bibinfo{person}{H. {Wang}},
  \bibinfo{person}{D. {Yershov}}, \bibinfo{person}{H. {Zhao}},
  \bibinfo{person}{M. {Benjamin}}, \bibinfo{person}{C. {Carr}},
  \bibinfo{person}{M. {Zuber}}, \bibinfo{person}{S. {Karaman}},
  \bibinfo{person}{E. {Frazzoli}}, \bibinfo{person}{D. {Del Vecchio}},
  \bibinfo{person}{D. {Rus}}, \bibinfo{person}{J. {How}}, \bibinfo{person}{J.
  {Leonard}}, {and} \bibinfo{person}{A. {Censi}}.}
  \bibinfo{year}{2017}\natexlab{}.
\newblock \showarticletitle{Duckietown: An open, inexpensive and flexible
  platform for autonomy education and research}. In
  \bibinfo{booktitle}{\emph{2017 IEEE International Conference on Robotics and
  Automation (ICRA)}}. \bibinfo{pages}{1497--1504}.
\newblock
\urldef\tempurl%
\url{https://doi.org/10.1109/ICRA.2017.7989179}
\showDOI{\tempurl}


\bibitem[\protect\citeauthoryear{Ramakrishna, Rahiminasab, Easwaran, and
  Dubey}{Ramakrishna et~al\mbox{.}}{2020}]%
        {ramakrishna2020efficient}
\bibfield{author}{\bibinfo{person}{Shreyas Ramakrishna}, \bibinfo{person}{Zahra
  Rahiminasab}, \bibinfo{person}{Arvind Easwaran}, {and}
  \bibinfo{person}{Abhishek Dubey}.} \bibinfo{year}{2020}\natexlab{}.
\newblock \showarticletitle{Efficient Multi-Class Out-of-Distribution Reasoning
  for Perception Based Networks: Work-in-Progress}. In
  \bibinfo{booktitle}{\emph{2020 International Conference on Embedded Software
  (EMSOFT)}}. IEEE, \bibinfo{pages}{40--42}.
\newblock


\bibitem[\protect\citeauthoryear{{Shi}, {Huang}, {Song}, {Wang}, {Lin}, and
  {Wu}}{{Shi} et~al\mbox{.}}{2020}]%
        {9259236}
\bibfield{author}{\bibinfo{person}{W. {Shi}}, \bibinfo{person}{G. {Huang}},
  \bibinfo{person}{S. {Song}}, \bibinfo{person}{Z. {Wang}}, \bibinfo{person}{T.
  {Lin}}, {and} \bibinfo{person}{C. {Wu}}.} \bibinfo{year}{2020}\natexlab{}.
\newblock \showarticletitle{Self-Supervised Discovering of Interpretable
  Features for Reinforcement Learning}.
\newblock \bibinfo{journal}{\emph{IEEE Transactions on Pattern Analysis and
  Machine Intelligence}} (\bibinfo{year}{2020}), \bibinfo{pages}{1--1}.
\newblock
\urldef\tempurl%
\url{https://doi.org/10.1109/TPAMI.2020.3037898}
\showDOI{\tempurl}


\bibitem[\protect\citeauthoryear{Tian}{Tian}{[n.d.]}]%
        {lanenavigation}
\bibfield{author}{\bibinfo{person}{David Tian}.}
  \bibinfo{year}{[n.d.]}\natexlab{}.
\newblock \bibinfo{booktitle}{\emph{DeepPiCar - Part 4: Autonomous Lane
  Navigation via OpenCV}}.
\newblock
\urldef\tempurl%
\url{https://towardsdatascience.com/deeppicar-part-4-lane-following-via-opencv-737dd9e47c96}
\showURL{%
\tempurl}


\bibitem[\protect\citeauthoryear{{Tim}, {Szemenyei}, and {Moni}}{{Tim}
  et~al\mbox{.}}{2020}]%
        {9263406}
\bibfield{author}{\bibinfo{person}{M. {Tim}}, \bibinfo{person}{M. {Szemenyei}},
  {and} \bibinfo{person}{R. {Moni}}.} \bibinfo{year}{2020}\natexlab{}.
\newblock \showarticletitle{Simulation to Real Domain Adaptation for Lane
  Segmentation}. In \bibinfo{booktitle}{\emph{2020 23rd International Symposium
  on Measurement and Control in Robotics (ISMCR)}}. \bibinfo{pages}{1--6}.
\newblock
\urldef\tempurl%
\url{https://doi.org/10.1109/ISMCR51255.2020.9263406}
\showDOI{\tempurl}


\bibitem[\protect\citeauthoryear{Varshney and Alemzadeh}{Varshney and
  Alemzadeh}{2017}]%
        {varshney2017}
\bibfield{author}{\bibinfo{person}{Kush~R Varshney} {and} \bibinfo{person}{Homa
  Alemzadeh}.} \bibinfo{year}{2017}\natexlab{}.
\newblock \showarticletitle{On the safety of machine learning: Cyber-physical
  systems, decision sciences, and data products}.
\newblock \bibinfo{journal}{\emph{Big data}} \bibinfo{volume}{5},
  \bibinfo{number}{3} (\bibinfo{year}{2017}), \bibinfo{pages}{246--255}.
\newblock


\bibitem[\protect\citeauthoryear{Vernekar, Gaurav, Abdelzad, Denouden, Salay,
  and Czarnecki}{Vernekar et~al\mbox{.}}{2019}]%
        {vernekar2019out}
\bibfield{author}{\bibinfo{person}{Sachin Vernekar}, \bibinfo{person}{Ashish
  Gaurav}, \bibinfo{person}{Vahdat Abdelzad}, \bibinfo{person}{Taylor
  Denouden}, \bibinfo{person}{Rick Salay}, {and} \bibinfo{person}{Krzysztof
  Czarnecki}.} \bibinfo{year}{2019}\natexlab{}.
\newblock \showarticletitle{Out-of-distribution detection in classifiers via
  generation}.
\newblock \bibinfo{journal}{\emph{arXiv preprint arXiv:1910.04241}}
  (\bibinfo{year}{2019}).
\newblock


\end{thebibliography}

\end{document}